\documentclass[journal,twocolumn]{IEEEtran}

\usepackage{cite}
\usepackage{graphicx}
\usepackage{multirow}
\usepackage{multicol}
\usepackage{amsmath}
\usepackage{amssymb}
\usepackage{array}

\usepackage{pifont}

\usepackage{hyperref}
\hypersetup{bookmarks=false,%
       bookmarksnumbered,%
           bookmarksopen,%
              colorlinks,%
              linkcolor = blue,
              plainpages=false,%
              breaklinks=true,%
            pdfstartview=FitH}

\ifCLASSINFOpdf
\else
\fi

\begin{document}

\title{Fast 3D Modeling of Anthropomorphic Robotic Hands Based on A Multi-layer Deformable Design}

\author{Li Tian\textsuperscript{1}, Hanhui Li\textsuperscript{1*}, Muhammad Faaiz Khan Bin Abdul Halil\textsuperscript{1}, Nadia Magnenat Thalmann\textsuperscript{1}, Daniel Thalmann\textsuperscript{2} and Jianmin Zheng\textsuperscript{1}
\thanks{
\newline
\textsuperscript{1}Nanyang Technological University, Singapore. \newline
\textsuperscript{2}École Polytechnique Fédérale de Lausanne, Switzerland. \newline
\textsuperscript{*}Hanhui Li is the corresponding author (hanhui.li@ntu.edu.sg).
}
}

\maketitle


\begin{abstract}
Current anthropomorphic robotic hands mainly focus on improving their dexterity by devising new mechanical structures and actuation systems. However, most of them rely on a single structure/system (e.g., bone-only) and ignore the fact that the human hand is composed of multiple functional structures (e.g., skin, bones, muscles, and tendons). This not only increases the difficulty of the design process but also lowers the robustness and flexibility of the fabricated hand. Besides, other factors like customization, the time and cost for production, and the degree of resemblance between human hands and robotic hands, remain omitted. To tackle these problems, this study proposes a 3D printable multi-layer design that models the hand with the layers of skin, tissues, and bones. The proposed design first obtains the 3D surface model of a target hand via 3D scanning, and then generates the 3D bone models from the surface model based on a fast template matching method. To overcome the disadvantage of the rigid bone layer in deformation, the tissue layer is introduced and represented by a concentric tube based structure, of which the deformability can be explicitly controlled by a parameter. Besides, a low-cost yet effective underactuated system is adopted to drive the fabricated hand. The proposed design is tested with 33 widely used object grasping types, as well as special objects like fragile silken tofu, and outperforms previous designs remarkably. With the proposed design, anthropomorphic robotic hands can be produced fast with low cost, and be customizable and deformable.
\end{abstract}  



\section{Introduction}

Designing anthropomorphic robotic hands that mimic the general aspects of the human hand is one of the most challenging problems in robotics, due to the sophisticated structure and functionality of the human hand \cite{gama2014anthropomorphic}. Ever since the early dexterous Stanford JPL hand \cite{Salisbury1983} was invented, there have been extensive inspiring advances \cite{rothling2007platform,schmitz2010design,deimel2016novel,controzzi2016sssa,piazza2017softhand,liu2020soft} in this area. Particularly, replacing rigid mechanical structures with soft materials, and simplifying the hand designs are emerging trends of research in recent years \cite{piazza2019century}.

From the perspective of human hand anatomy, most current robotic hands can be roughly divided into two groups: (i) Rigid-body robotic hands (e.g., bone based) \cite{rothling2007platform,gosselin2008anthropomorphic,deshpande2011mechanisms,xu2016design,faudzi2017index,tasi2019design} that consist of rigid, individual structures replicating phalanges and joints, such as Shadow Dexterous Hand \cite{rothling2007platform} and ACB hand \cite{tasi2019design}. (ii) Soft, tissue-like hands \cite{deimel2016novel,heung2019robotic,yamaguchi2019human,oh2020untethered,yang2020hybrid,hashemi2020bone} that are built using soft materials and actuators, among which RBO hand 2 \cite{deimel2016novel} is the typical example. These two types of robotic hands have their own advantages and disadvantages: rigid-body hands are reliable, sturdy, and better reflect the movement kinematics and dynamics of the human hand, but they require sophisticated mechanisms to perform dexterous operations. In contrast, soft robotic hands are deformable, inherently safer, and easier to control. However, for many human-centered applications like social robotics, service robotics, and entertainment, the degree of resemblance with the human hand is critical \cite{bainbridge2004berkshire}. Hence soft robotic hands are not optimal for these applications. Besides, neither of these two types is easy to customize, which means they are built with predefined shapes and sizes, instead of a target given on-the-fly.

This study aims at combining the advantages of both rigid and soft robotic hands, to fabricate deformable, cost-effective robotic hands with the customizable human-like appearance. To this end, we notice that the disadvantages of previous methods are mainly caused by \emph{the over-simplification of the human-hand structure}, which assumes that the functionality of human hand can be realized by a single structure. Such an assumption actually increases the complexity of the design process and limits the candidate materials for fabrication, as various mechanical structures and actuators need to be compatible and integrated. Therefore, instead of relying on a single structure, we propose a novel multi-layer design, which highly replicates the anatomy of the human hand. Our design is composed of three layers, namely, \emph{skin, tissue and bone}, and each of them is devised for a particular purpose: the skin layer provides the human-like appearance; the tissue layer is made of soft materials to guarantee the deformability of the hand; the bone layer serves as the rigid, supportive structure and is attached to actuators to enable object grasping.

To ensure the proposed design can be fabricated efficiently, we exploit the advantages of 3D scanning and 3D printing in rapid and adaptable fabrication \cite{rus2015design,yap2016high,rich2018untethered}. A fast template matching approach and a concentric tube based structure are proposed to obtain the 3D models of bones and tissues, respectively. Besides, an underactuated system, which simplifies previous complex motion control systems \cite{faudzi2017index,tian2019design2}, is introduced in this paper.

Table \ref{tab: design} summarizes the differences between the proposed design and other robotic hands. To the best of our knowledge, this paper proposes the first customizable multi-layer design for anthropomorphic robotic hands. The proposed design allows us to fully take the advantage of soft robotic hands in deformation. We conduct extensive experiments, including the standard Feix object grasping test \cite{feix2015grasp} and the trajectory test, to demonstrate that our design can effectively simulate the functionality of the human hand. Due to the deformability of our design, special objects, including extremely fragile silken tofu and smooth marbles, can be grasped by our fabricated hand. In summary, compared with previous single layer/structure hands, the proposed one is not only simpler and easier to fabricate, but also more adaptive and robust.    

\begin{table*}[!t]
\centering
\caption{Comparison of the proposed design with other anthropomorphic robotic hands. Materials and E-motor stand for 3D printing materials and electric motor, respectively.}
\label{tab: design}
\footnotesize
\begin{tabular}[c]{|*{7}{c|}}
\hline
\multirow{2}{*}{{Model}} & \multirow{2}{*}{{Year}} & \multicolumn{3}{c|}{{Materials and Components}} & \multirow{2}{*}{{Actuator}} & \multirow{2}{*}{{Joint Type}} \\ \cline{3-5} & & Skin & Tissue & Bone & & \\ \hline
Shadow hand \cite{rothling2007platform} & 2007 & N.A. & \multicolumn{2}{c|}{Rigid structures} & E-motor + tendon & Rigid \\ \hline
DEXMART hand \cite{palli2014dexmart} & 2014 & N.A. & \multicolumn{2}{c|}{Rigid structures} & E-motor + tendon & Dislocatable \\ \hline
RBO hand V2 \cite{deimel2016novel} & 2015 & N.A. & \multicolumn{2}{c|}{Soft materials} & Pneumatic motor & Soft continuous \\ \hline
Soft robotic hand \cite{she2015design} & 2015 & N.A. & \multicolumn{2}{c|}{Soft materials} & Shape memory alloy & Soft continuous \\ \hline
Biomimetic hand \cite{xu2016design} & 2016 & N.A. & N.A. & Rigid materials & E-motor + tendon & Flexible \\ \hline
ACB hand \cite{tasi2019design} & 2019 & N.A. & N.A. & PolyJet, resin & E-motor + tendon & Flexible \\ \hline
Nadine hand V4 \cite{tian2019design} & 2019 & Silicone rubber & \multicolumn{2}{c|}{Flexible materials} & E-motor + tendon & Rigid \\ \hline
Ours & 2020 & Silicone rubber & Elastic materials & Rigid materials & E-motor + tendon & Flexible \\ \hline
\end{tabular}
\end{table*}

\section{Materials and Methods}

\subsection{The multi-layer deformable design}
The goals of our anthropomorphic robotic hands can be specified into two aspects: first, the modeling process of our robotic hands should be fast and customizable, so that we can fabricate hands of various shapes and sizes easily; second, our robotic hands should have the similar functionality of the human hand, especially in completing different grasping gestures. To this end, we propose a multi-layer deformable design of anthropomorphic robotic hands, which mainly consists of (i) a layer of silicone skin, (ii) a layer of 3D printed tissues, (iii) a layer of 3D printed bones, and (iv) an underactuated system.

Specifically, given a real human hand as the target, we first obtain its surface model via a 3D scanner. The skin layer could be de-molded directly from the target hand to mimic the appearance of the target hand. And then we propose a fast template matching method to obtain the corresponding 3D bone models based on the surface model (Section \ref{sec: bone}). After that, an effective concentric tube structure is adopted to construct the tissue layer (Section \ref{sec: tissue}). The tissue layer serves as the intermediate layer between the skin and bones, and is made of elastic materials to allow the high deformability of our robotic hands. At last, we adopt a low-cost cable driven system, to provide our robotic hands with the mobility and stability (Section \ref{sec: joint}).

The visualization of the above design and fabrication process can be found in \footnote{\url{https://entuedu-my.sharepoint.com/:v:/g/personal/hanhui_li_staff_main_ntu_edu_sg/EWukvnt4Mh1IjH_xNH_eXX0BJ2GF8xMpm4gXmS3iLsIwxA?e=Bw9itD}}. Compared with conventional single layer/structure robotic hands \cite{rothling2007platform,xu2016design,tasi2019design,tian2019design,tian2018methodology}, our multi-layer design is more similar to human hands from the perspective of biomimetics. More importantly, as we will demonstrate in the experiment section, such a deformable design is versatile for grasping objects of different shapes, textures and materials.

\subsection{Materials}

For the purpose of easy customization and fabrication, all materials used in this paper are low-cost and widely accessible. Based on Young's modulus \cite{rus2015design}, we consider two 3D printing materials that have the desirable tensile strength and are compatible with the Form-2 3D printer. For the bone models, we use \emph{rs-f2-gpcl-04}, which is a rigid material and has 2.8 GPa tensile strength. For the tissue layer, we use an elastic 3D printable material, \emph{rs-f2-elcl-01}, which has 50A shore hardness and 3.23 MPa tensile strength. The skin layer is made of silicone rubber and we choose \emph{PlatSil Gel 10} with 10A shore hardness. 

\begin{figure}[!t]
    \centering
    \includegraphics[width = 0.5\textwidth]{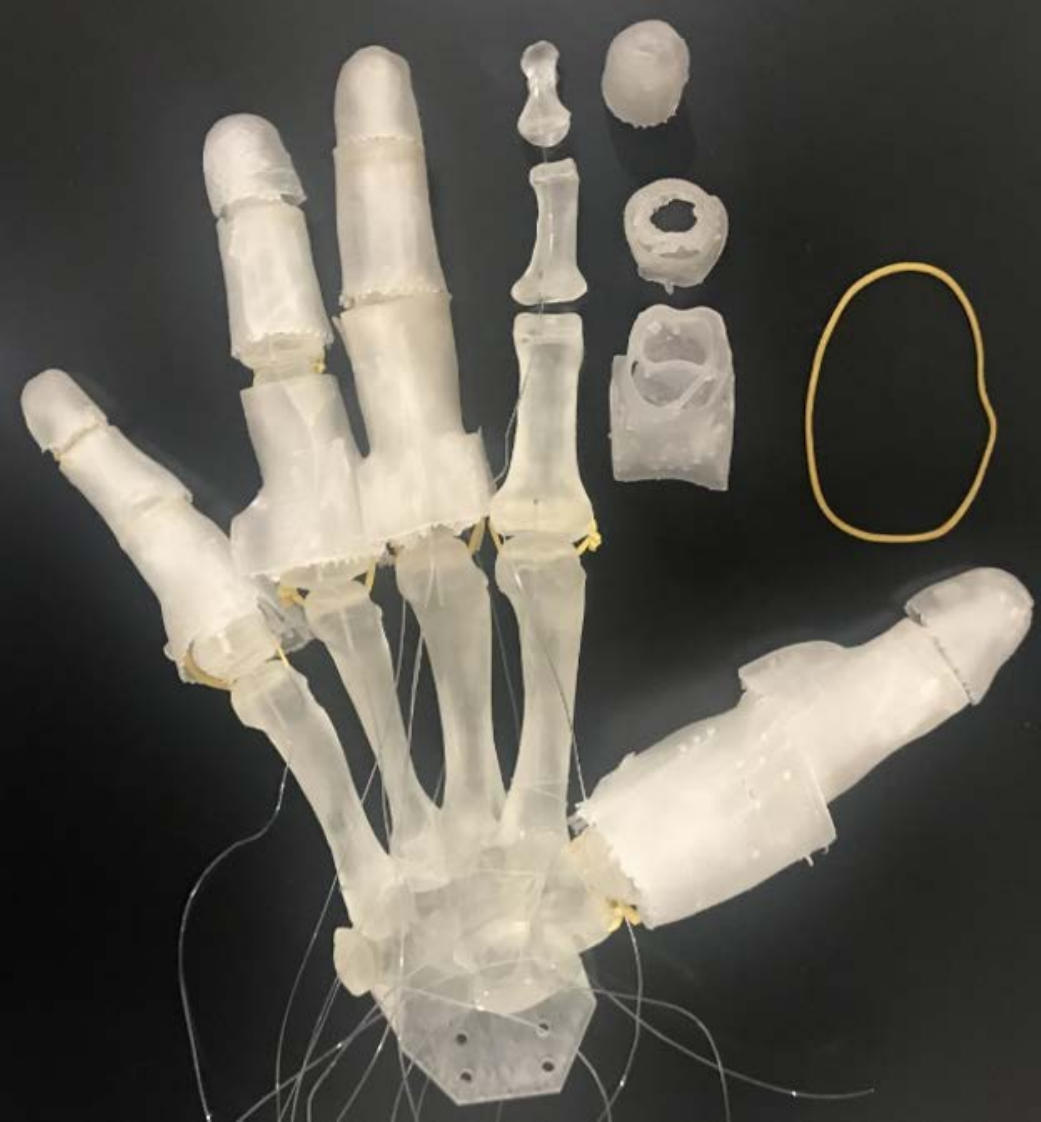}
\caption{Major materials and components of our robotic hands. Soft and stiff 3D printing materials are used for producing the tissues and bones, respectively. Nylon cables and rubber bands are used for simulating tendons and ligaments.}
\label{fig: material}
\end{figure}

As to our actuation system, we use rubber bands as ligaments to connect bones. Nylon cables of 0.3 mm diameter and 10 lbs average breaking force are used as tendons. The servo motor, \emph{HITEC HS-5070HM}, which is light (12.7 g) yet with high torque (3.6 kg/cm), is used for driving the tendons. Fig. \ref{fig: material} demonstrates the major materials and components of a fabricated robotic hand based on our design.

\subsection{3D modeling of bones}
\label{sec: bone}

Modeling bones is the necessary procedure of customizing a highly biomimetic robotic hand, because it provides not only the basic structure of the hand but also the perfect guide for hand motions. However, the direct modeling of bones is difficult, as they are hidden beneath the skin. Radiology methods, such as computed tomography and magnetic resonance imaging, are costly for the customized fabrication. Therefore, we propose to first obtain the surface model of the target hand via 3D scanning, and then generate the 3D mesh models of bones based on our fast template matching method. 

\begin{figure*}[!t]
    \centering
    \includegraphics[width = 1\textwidth]{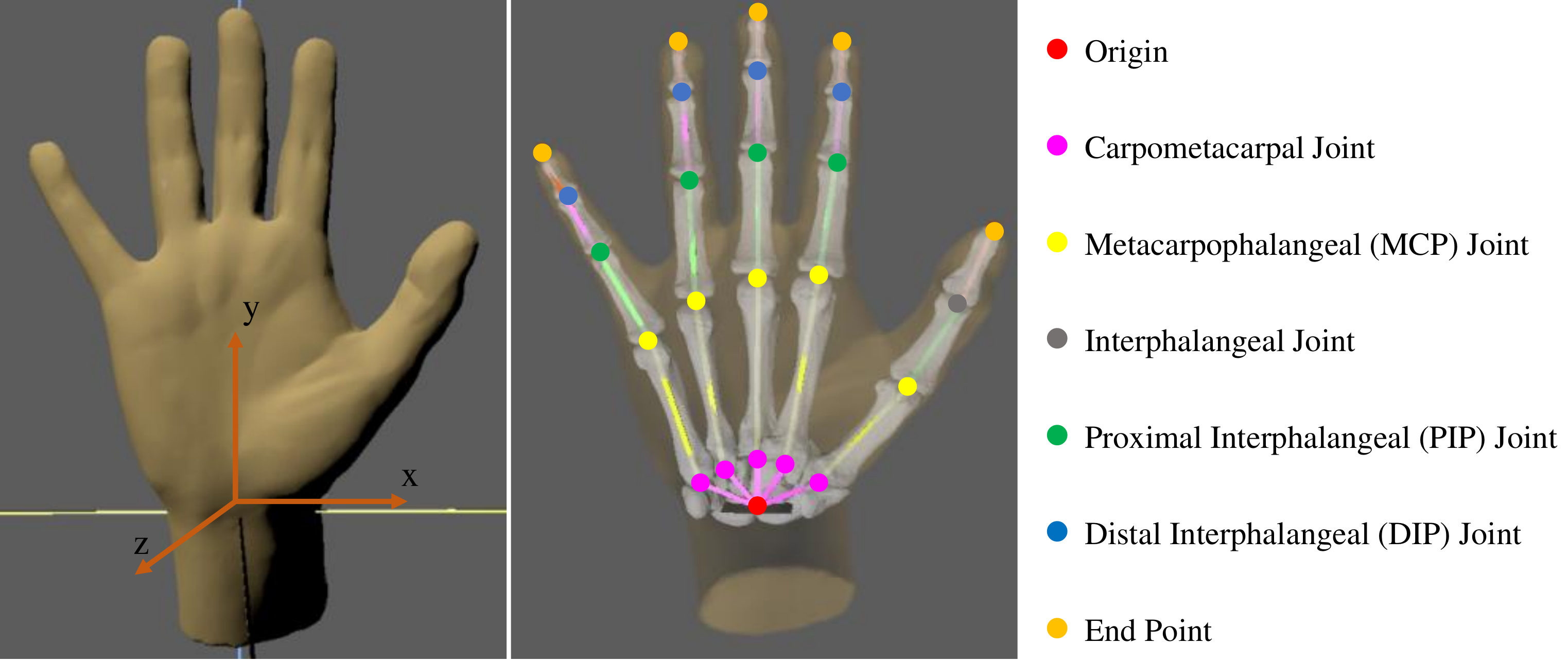}
\caption{Left: The surface model captured by 3D scanning. Right: The anatomy of the human hand. 25 landmarks are labeled to locate joints and bones.}
\label{fig: bone}
\end{figure*}

3D scanning provides a faster and more accurate way of modeling, compared with traditional modeling techniques \cite{shah2017review}. Our earlier study \cite{tian2018methodology} also demonstrates that it is possible to maintain more than $90\%$ of the geometric information of a human hand via 3D scanning. Thus, given a target hand, we scan it with the \emph{Go! Scan 50} 3D scanner, to obtain the corresponding triangle mesh based 3D model. As shown in Fig. \ref{fig: bone}, the scanned 3D model is a vivid and precise representation of the surface of our target.

Next, we propose to estimate a series of geometric transformations, to match a template of 3D bone models to the surface model. We place landmarks on the surface model to determine the positions, orientations and sizes of bones. As the bones have relatively fixed proportions and connections \cite{buryanov2010proportions,li2008validation}, we adjust the surface model to make it be symmetric w.r.t. the $xy$-plane (Fig. \ref{fig: bone}), so that the $z$ coordinates can be ignored and we can use 2D landmarks to reduce the complexity of labeling and computation. Based on the anatomy of the human hand, 25 landmarks are used in total, as shown in Fig. \ref{fig: bone}. In this way, each bone is uniquely determined by two landmarks, and we can construct a local 2D coordinate system for each bone, in which one of the landmarks is considered as the origin, while the other one is used as the reference to estimate the transformations. That is, let $P$ denote the point set of one of the template bone models, and $\mathbf{p} = [p_x, p_y]^{T} \in P$ be an arbitrary point, the corresponding coordinates of $\mathbf{p}$ in the local coordinate system of the customized bone model, $\mathbf{p}'$, is calculated as follows: 
\begin{equation}
\mathbf{p}'=R(\theta)S(\lambda) \mathbf{p},
\end{equation}
where $R(\theta)$ is the rotation matrix parametrized by $\theta$, the counterclockwise angle of rotation w.r.t. the $x$ axis:
\begin{equation}
R(\theta ) = \left[ {\begin{array}{*{20}{c}}
  {\cos \theta }&{ - \sin \theta } \\ 
  {\sin \theta }&{\cos \theta } 
\end{array}} \right].
\end{equation}
And $S(\lambda)$ is the uniform scaling matrix with the scaling factor $\lambda$: 
\begin{equation}
S(\lambda ) = \left[ {\begin{array}{*{20}{c}}
  \lambda &0 \\ 
  0&\lambda  
\end{array}} \right].
\end{equation}
$\theta$ and $\lambda$ can be determined by the reference landmark. Let $\mathbf{r}'$ and $\mathbf{r}$ denote the reference landmark in the customized and the template local coordinate system, respectively. For $\theta$, we estimate it via the inverse trigonometric functions like $arcsine$, while for $\lambda$, we consider $\lambda = \frac{{\left\| \mathbf{r}' \right\|_2}}{{\left\| \mathbf{r} \right\|_2}}$, where ${{\left\| \cdot \right\|_2}}$ is the 2-norm.

Before 3D printing the bones, we also put holes at the ends of them, so that we can easily connect the bones with rubber bands (as ligaments). Unlike the previous method \cite{tian2018methodology} that requires the specific transformations for different landmarks, the above matching process can be applied to all 3D bone models of the template to generate the customized models. This improves the efficiency of producing customized robotic hands significantly, as the only manual operation is to locate the 2D landmarks, which can be completed within a few minutes. 

\subsection{3D modeling of tissue}
\label{sec: tissue}

The tissue layer is critical to our anthropomorphic hands, as it helps to overcome the disadvantages of rigid, bone-only hands in object grasping. From the anatomical perspective, the soft tissues surrounding the fingers are of complex types and structures, such as subcutaneous fat and muscles. Hence, the core of modeling the tissue layer is to design a unified representational structure, which is 3D printable and deformable.

\begin{figure}[!t]
    \centering
    \includegraphics[width = 0.5\textwidth]{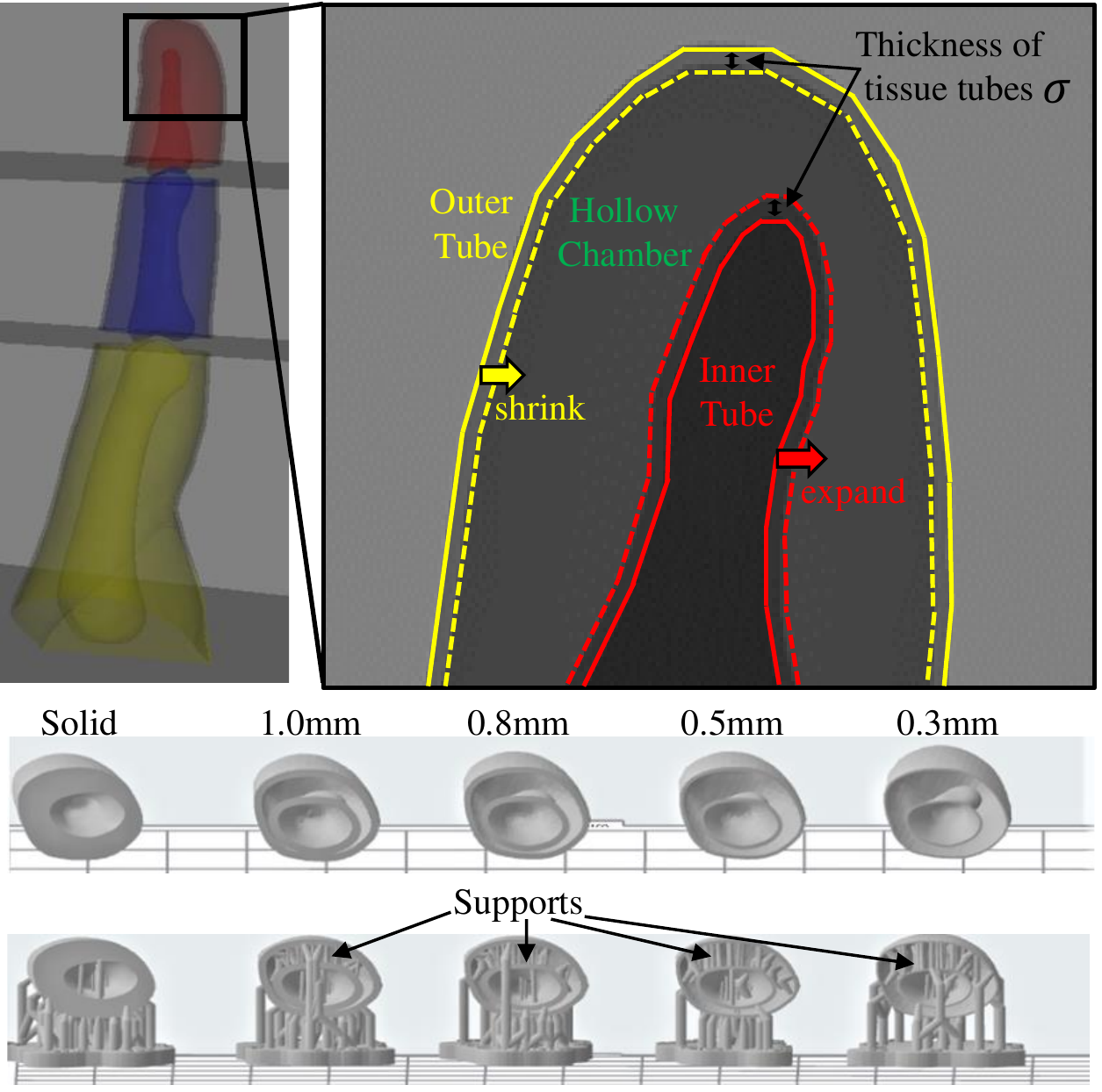}
\caption{Top: Modeling the tissue layer as a concentric tube structure. We shrink the surface model (marked with the yellow solid curve) while expand the bone layer (marked with the red solid curve) to obtain the two tubes of the tissue layer. Middle: Demonstration of tissue tubes with different thicknesses. Bottom: Supports for stabilizing the structure.}
\label{fig: tissue}
\end{figure}

Our solution for the tissue modeling problem is a novel concentric tube structure, which is constructed via the following three steps:

(i) {Hull generation}. As the tissue layer cannot be scanned and modeled directly, we propose to generate a 3D hull to determine the surfaces of the tissue layer. Note that we already obtain the 3D mesh models of the surface and bones via the fast template matching method, therefore, we consider the surface model as the outer hull, while the bone model as the inner hull, as demonstrated in Fig. \ref{fig: tissue}. This allows us to have a basic structure to model and 3D print the tissue layer.

(ii) {Structure hollowing}. This step aims at providing the basic structure with the deformability. Although the material used for the tissue layer (\emph{rs-f2-elcl-01}) is a common choice for soft structures, it is much harder than the finger tissues. Therefore, instead of using a solid structure, we hollow the model of the tissue layer and add supports, to obtain a more elastic tube shaped structure. Any support structure could be used and we simply use the default one provided by the Form-2 3D printer. This also brings a side benefit that we can reduce the cost of 3D printing materials, e.g., the solid design of the tissue surrounding the distal phalanx of the index finger requires about 2.9 ml material, while the hollow one only requires about 1.4 ml.

(iii) {Deformation curve fitting}. We further introduce an extra parameter $\sigma$, i.e., the thickness of the tissue tube, to control the deformation modulus of our tissue layer. This is feasible as the larger $\sigma$ is, the harder it is to deform the tissue layer. And the 3D tissue model with the thickness of $\sigma$ can be generated easily by shrinking the outer hull by $\sigma$, while expanding the inner hull by $\sigma$ as well. In this way, the tissue layer is modeled by two concentric tubes, as demonstrated in Fig. \ref{fig: tissue}. To determine the optimal value of $\sigma$, inspired by the Young's modulus, we record the curves of tensile forces and axial strains with different values of $\sigma$, and set $\sigma$ as the one with the curve being closest to that of the human finger.

\subsection{The underactuated system}
\label{sec: joint}

In this section, we describe the proposed underactuated system which simulates joints, ligaments, tendons and muscles of the human hand. Ligaments are fibrous connective tissues that connect bones to other bones and form joints. Tendons are connective tissues as well, but they are attached to muscles and bones. To mimic the dynamics of human hands, we use cables as tendons, with electric servo motors as muscles to control the motions of the finger.

\begin{figure}[!t]
    \centering
    \includegraphics[width = 0.5\textwidth]{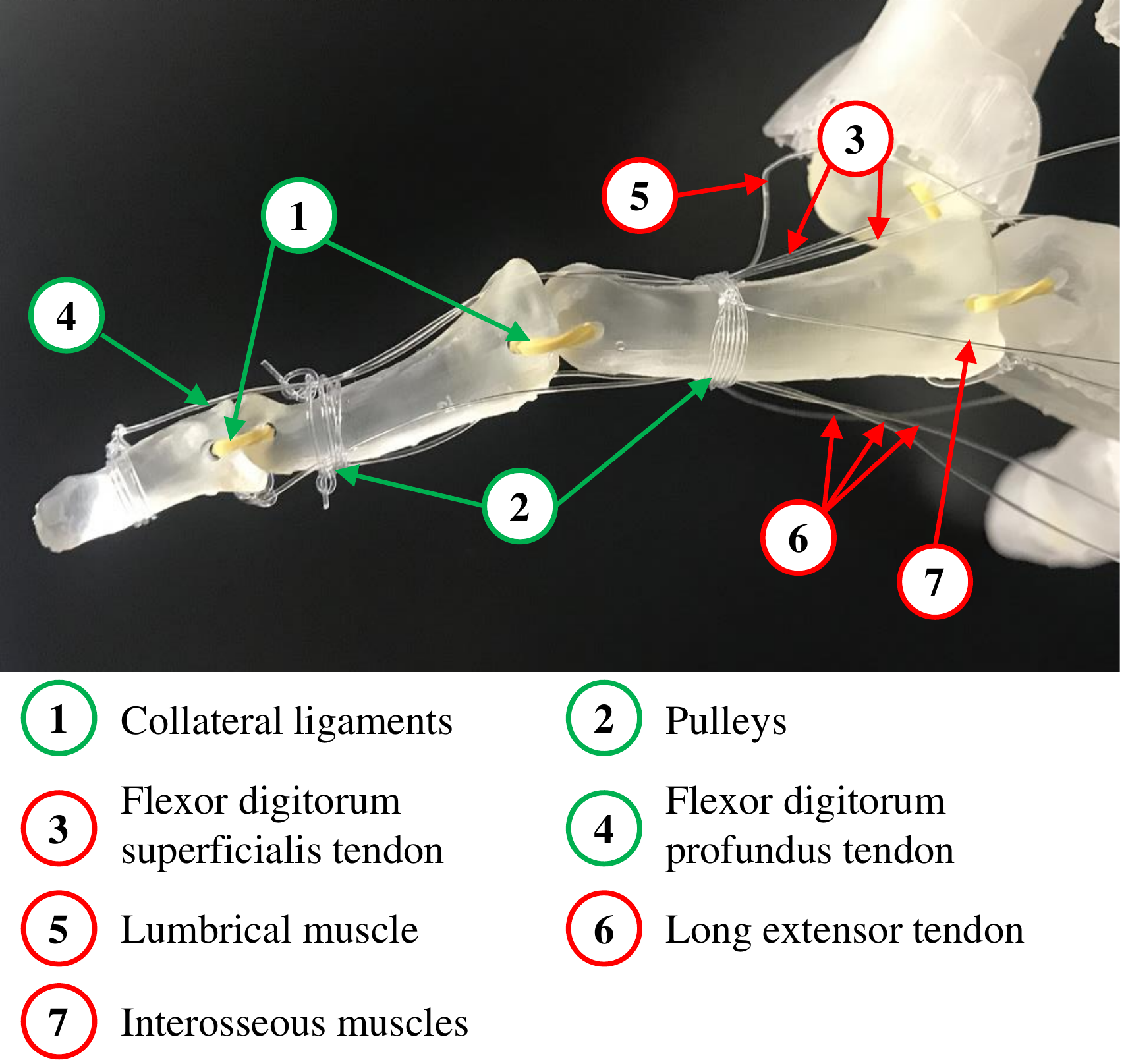}
\caption{Demonstration of the underactuated system of our robotic hands. Circle 1-7 denote structures used in our previous method \cite{tian2019design2}. With the proposed multi-layer design, the system is simplified (remaining structures are labeled in green) to reduce the complexity and cost.}
\label{fig: ligament}
\end{figure}

To begin with, we introduce Lansmeer model III \cite{landsmeer1961studies}, which describes the movement of a tendon and can be approximated via a second order polynomial \cite{buchner1988dynamic,brook1995biomechanical} as follows:
\begin{equation}
L = (b + h\phi )\phi,
\label{eq:tendon}
\end{equation}
where $L$ is the displacement of the tendon, $\phi$ is the joint angle, $b$ and $h$ are empirical constants. To realize the fine control of the finger, our previous design \cite{tian2019design2} further extends Eq. (\ref{eq:tendon}) to construct a 3-stage cable driven system as follows:
\begin{equation}
\begin{split}
  {L_p} = & ({b_p} + {h_p}{\phi _p}){\phi _p} \hfill, \\
  {L_i} = & {L_p} + ({b_i} + {h_i}{\phi _i}){\phi _i} \hfill, \\
  {L_d} = & {L_p} + {L_i} + ({b_d} + {h_d}{\phi _d}){\phi _d} \hfill, \\ 
\end{split}
\end{equation}
where indexes $p$, $i$ and $d$ denote variables related to the proximal, intermediate and distal phalanx, respectively. That is, each of these three phalanges is driven by a cable, and all three cables cooperate in completing the movement. Note that there are two groups of tendons, i.e., flexor tendons that bend the finger, and extensor tendons that straighten the finger. Consequently, such a design requires two groups of the driven system: one for the flexor tendons, including the flexor digitorum superficialis (FDS) tendon and the flexor digitorum profundus (FDP) tendon; while the other one for the long extensor tendon, as shown in Fig. \ref{fig: ligament}. With the other two driven cables for simulating the lumbrical and interosseous muscles, 8 driven cables and 4 motors (one motor can be shared by two cables) are used in total to control a finger. 
 
In this paper, due to our multi-layer design, the aforementioned control system can be simplified significantly. As the layers of skin and tissue are made of elastic materials, they are able to return to their original shapes automatically after being flexed, and consequently the bones are pulled back to their original positions as well. Thus, we remove the cables of the extensor tendon, the lumbrical and interosseous muscles. Furthermore, for the flexor tendons, we only keep the cable attached to the FDP tendon. As shown in Fig. \ref{fig: ligament}, the new 
underactuated system requires only one cable and one motor for each finger (except the thumb which uses two cables and two motors). This reduces the complexity and size of our system remarkably, and hence our system can be easily installed on various robotic arms.
 
\section{Experiments and Results}

\subsection{Object grasping test}

We first evaluate the performance of the proposed design in object grasping. Following previous methods \cite{deimel2016novel,xu2016design,tasi2019design,tian2019design}, we consider the grasp taxonomy defined by Feix et al. \cite{feix2015grasp}, which consists of 33 human grasp types. In our experiments, a successful grasp is an every static one-hand posture with which an object can be held securely. Each grasp type is tested 10 times and we use the success rate as the evaluation metric. Two state-of-the-art robotic hands, including InMoov hand \cite{langevin2014inmoov} and Nadine's hand V4 \cite{tian2019design}, are fabricated for comparison, because the models and hardware specifications of InMoov hand are publicly available, while Nadine's hand V4 is similar to our design.

\begin{figure}[!t]
    \centering
    \includegraphics[width = 0.5\textwidth]{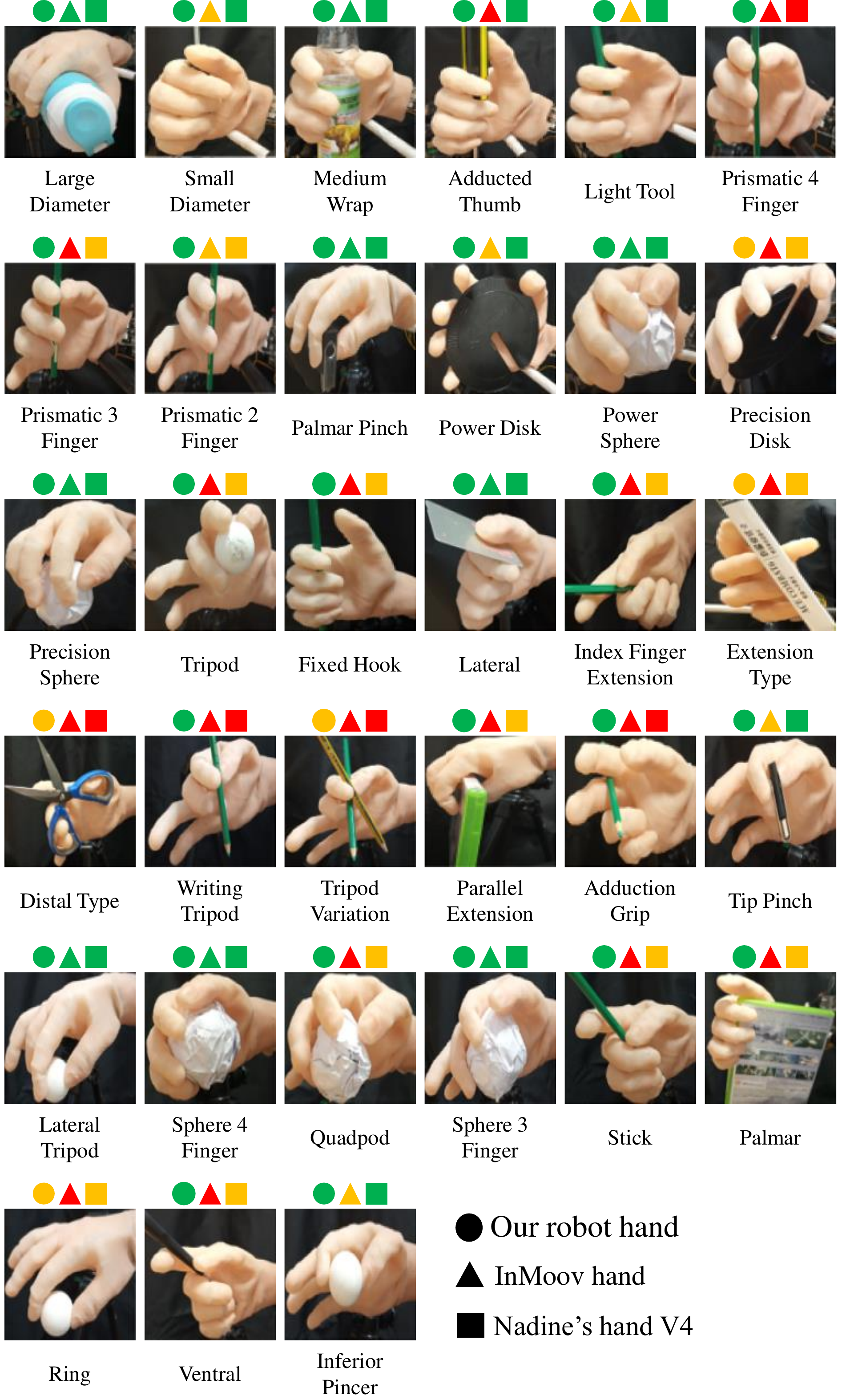}
\caption{Comparison of object grasping success rates of different methods. Methods with success rates $\ge 0.8$, in $[0.2, 0.8)$ and $< 0.2$ are labeled in green, yellow and red, respectively. The proposed robotic hand has the highest success rates for all grasping gestures.}
\label{fig: sota}
\end{figure}

The results of the object grasping test are reported in Fig. \ref{fig: sota}. These results show that the proposed robotic hand completes most grasp types with high success rates ($\ge 80\%$). Furthermore, the performance of our robotic hand is superior to that of InMoov hand and Nadine's hand V4 with all 33 grasp types. Note that the number of actuators of our robotic hand is equal to that of Nadine's hand V4 (i.e., 6 actuators), which indicates that the success of our robotic hand mainly comes from the deformable multi-layer design.

\begin{figure}[!t]
    \centering
    \includegraphics[width = 0.5\textwidth]{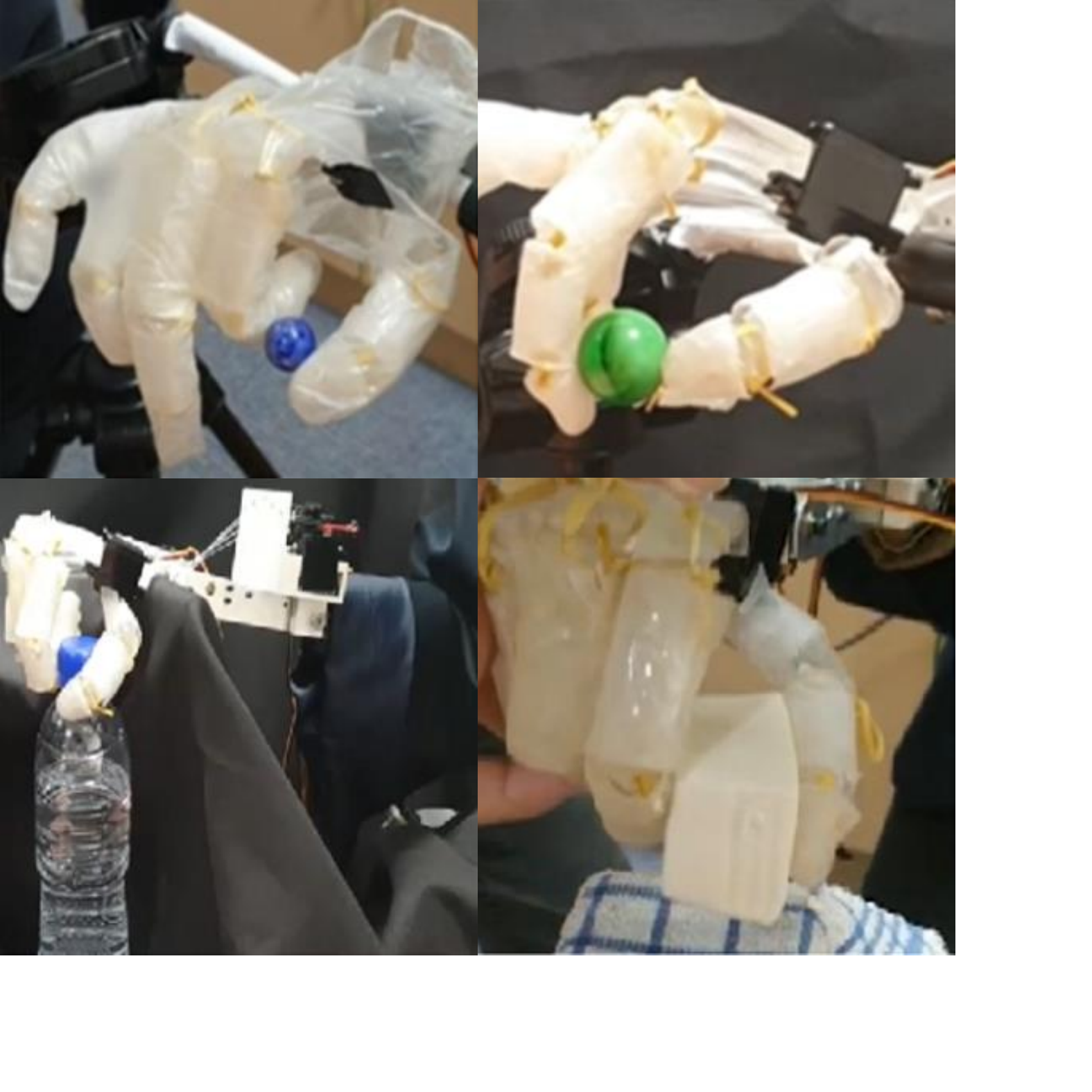}
\caption{Demonstration of our robotic hand in pinching special objects. Top row: marble with diameter of 16 mm (left) and 26 mm (right); Bottom left: a water bottle weighing 450 g. Bottom right: a block of silken (soft) tofu.}
\label{fig: specialobject}
\end{figure}

As we have emphasized, the deformability of robotic hand is the key of grasping objects of various textures and weights. To validate this, we also conduct a series of experiments on grasping special objects. Fig. \ref{fig: specialobject} shows a few example of pinching special objects: our robotic hand successfully pinches smooth marbles with different diameters. It can also hold a water bottle weighing 450 g, while the maximum weight that InMoov hand can hold is less than 200 g. Even soft and extremely fragile objects like the silken tofu block can be picked up by our robotic hand as well. For more examples of grasping special objects, interested readers can refer to \footnote{\url{https://entuedu-my.sharepoint.com/:v:/g/personal/hanhui_li_staff_main_ntu_edu_sg/Ed3rbYf3KQNEkqAOyDYOyPEBM5RWBgMCAoFySvy2kqRp7g?e=pLsEHB}}.

\subsection{Ablation study of the multi-layer design}
To provide an insight into the effectiveness of the multi-layer design, we further conduct an ablation study in which our robotic hands with different layers are evaluated. Specifically, we perform the object grasping test with 3 variants of our design, including a \emph{Bone-only} design, a \emph{Bone + tissue} design and a \emph{Bone + skin} design. 

\begin{figure}[!t]
    \centering
    \includegraphics[width = 0.5\textwidth]{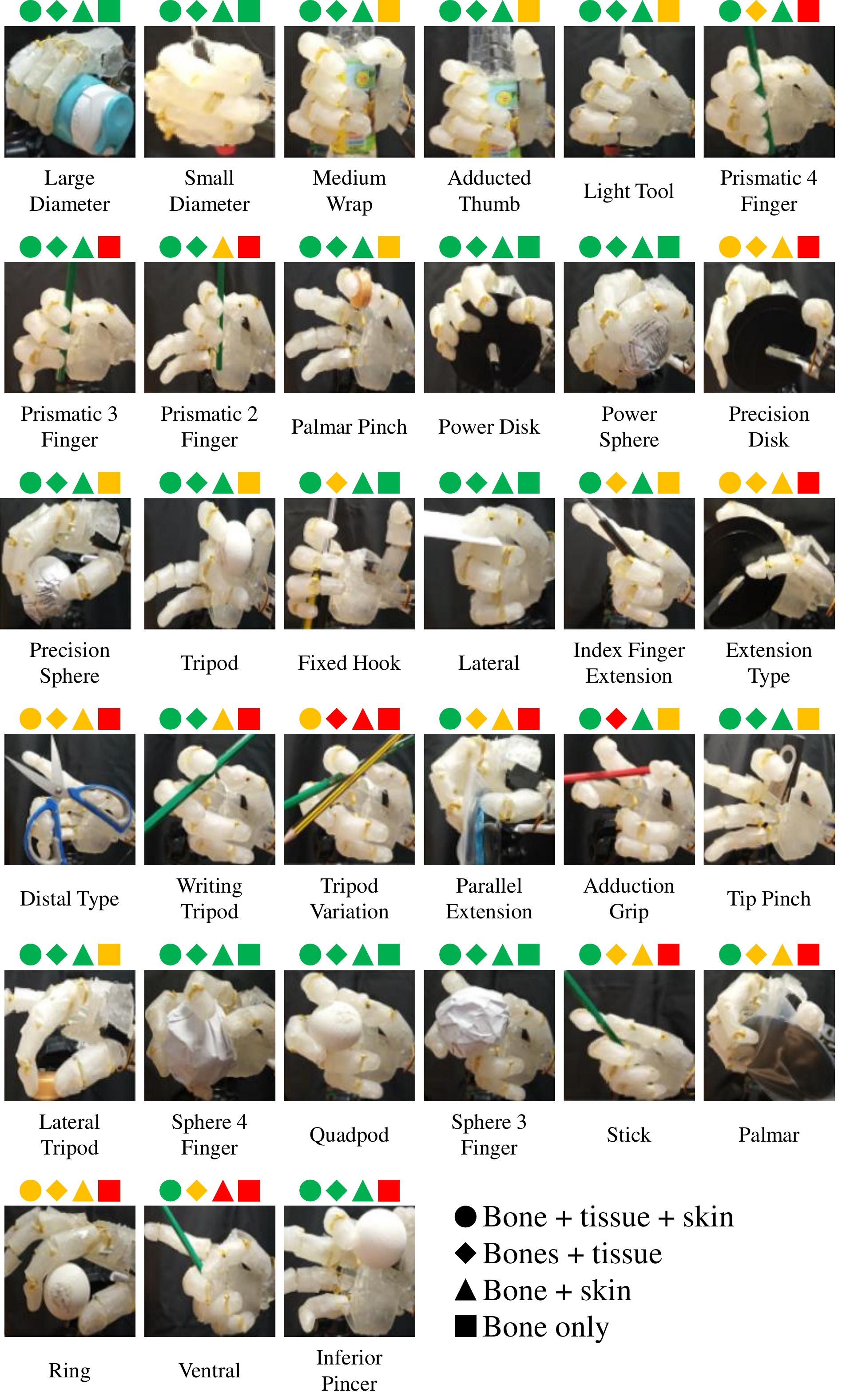}
\caption{Comparison of object grasping success rates of our robotic hands with different layers. Methods with success rates $\ge 0.8$, in $[0.2, 0.8)$ and $< 0.2$ are labeled in green, yellow and red, respectively. This clearly validates the effectiveness of each layer in our design.}
\label{fig: ablation}
\end{figure}

Fig. \ref{fig: ablation} demonstrates the results of the ablation study. We can obverse that the \emph{Bone-only} design fails with multiple grasp types, such as prismatic finger, extension and writing tripod. This again validates our opinion that single layer/structure robotic hands are hard to fully realize the ability of human hands. Nevertheless, with an extra layer of skin or tissue, we obtain the significant performance gains, e.g., the success rate of the ``prismatic 3 finger'' type rises from below $20 \%$ to higher than $80 \%$. And our design with all three layers obtains the highest success rates. Based on these results, we can conclude that our multi-layer design is a practicable solution for anthropomorphic robotic hands.

\subsection{Deformability of the tissue layer}

The deformability of our robotic hands is determined by the tissue layer. To demonstrate that our proposed concentric tube structure for the tissue layer can effectively simulate the deformability of the human hand, here we adopt the deformation curve \cite{shimawaki2007quasi} as the measure of deformability, and compare our structure with other methods.

\begin{figure*}[!t]
    \centering
    \includegraphics[width = 1\textwidth]{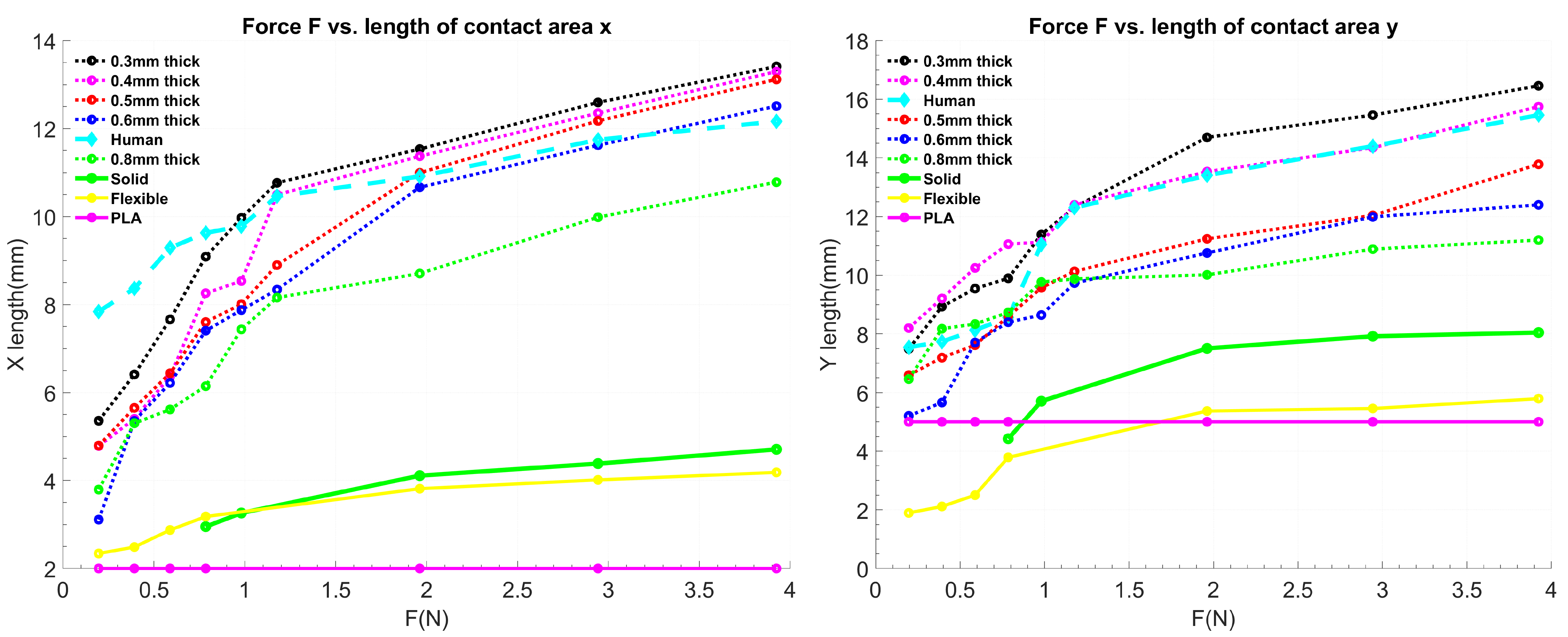}
\caption{Deformation curves of our tissue layer with different tube thicknesses. Compared with other materials and designs, our tissue layer with 0.4 mm thickness tubes better fits the deformation curves of the human finger.}
\label{fig: deform_tissue}
\end{figure*}

First, we consider two 3D printing materials for comparison, namely, the flexible material which is used by Nadine's hand V4, and Polylactic Acid (PLA) which is used by InMoov hand. As demonstrated in Fig. \ref{fig: deform_tissue}, the deformation curves of these two materials are far away from those of the human finger. As we have mentioned above, the soft material we used is still harder than human tissues, and hence merely using this material with the solid structure cannot achieve the desirable deformability. On the other hand, with our tube based structure, the deformability of the tissue layer is improved greatly.

More importantly, our experimental results suggest that our design can explicitly control its deformability via setting $\sigma$, the thickness of tubes. We report the deformation curves of $\sigma = \{0.3, 0.4, 0.5, 06, 0.8\}$ in Fig. \ref{fig: deform_tissue}. It can be observed that the deformability of our robotic hands is inversely proportional to $\sigma$. With $\sigma = 0.4$, we can obtain the tissue layer which has the deformation curves approximating those of the human finger.

\subsection{Trajectory Analysis}
\begin{figure}[!t]
    \centering
    \includegraphics[width = 0.5\textwidth]{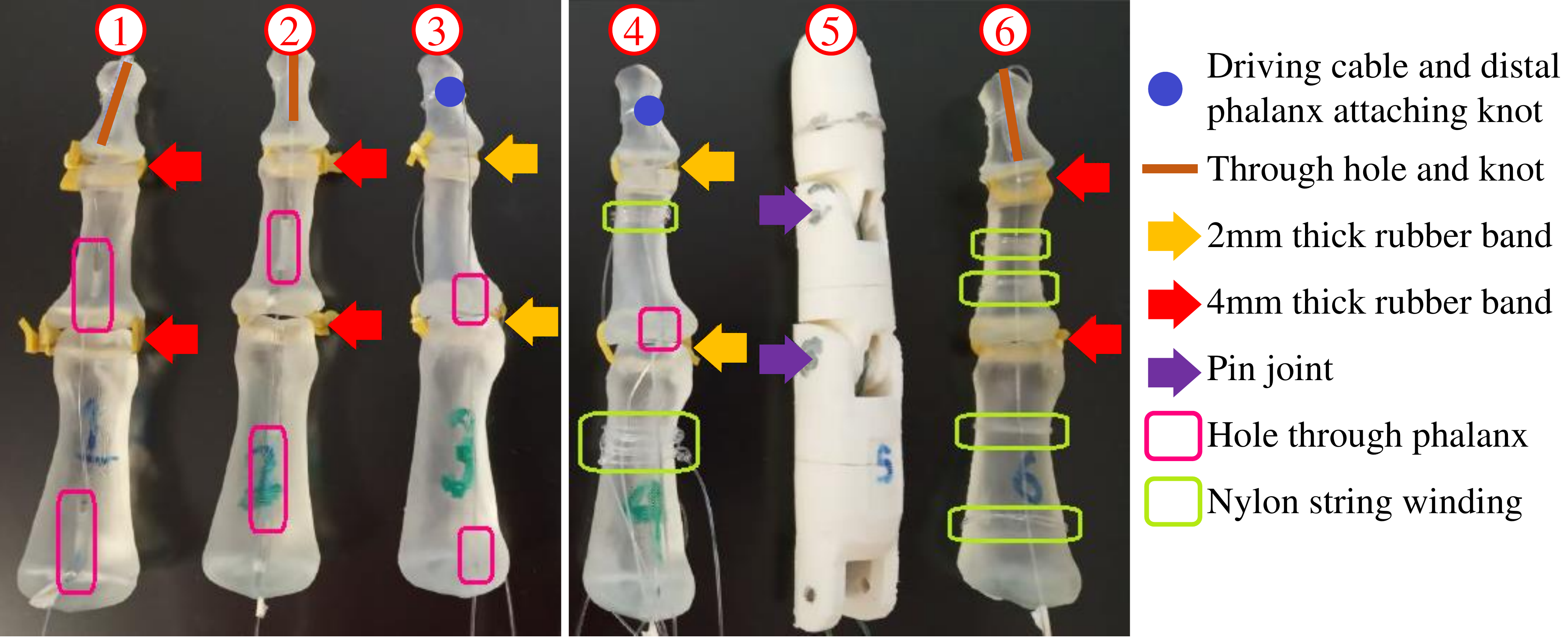}
\caption{Variants of the proposed underactuated system. Design 5 is adopted from InMoov hand. Design 6 is the optimal one and is used throughout this paper.}
\label{fig: finger}
\end{figure}

Last but not least, we propose to analyze the functionality of the underactuated system. As the system is responsible for flexion and extension, we record the trajectories of fingertips during flexion, to investigate whether they can fit the trajectory of the human finger. Our system can be implemented with different configurations, e.g., with different attaching knots and displacements of strings. Therefore, as demonstrated in Fig. \ref{fig: finger}, we realize 5 different designs of the system, and include the design from InMoov hand as the baseline for comparison.

\begin{figure}[!t]
    \centering
    \includegraphics[width = 0.5\textwidth]{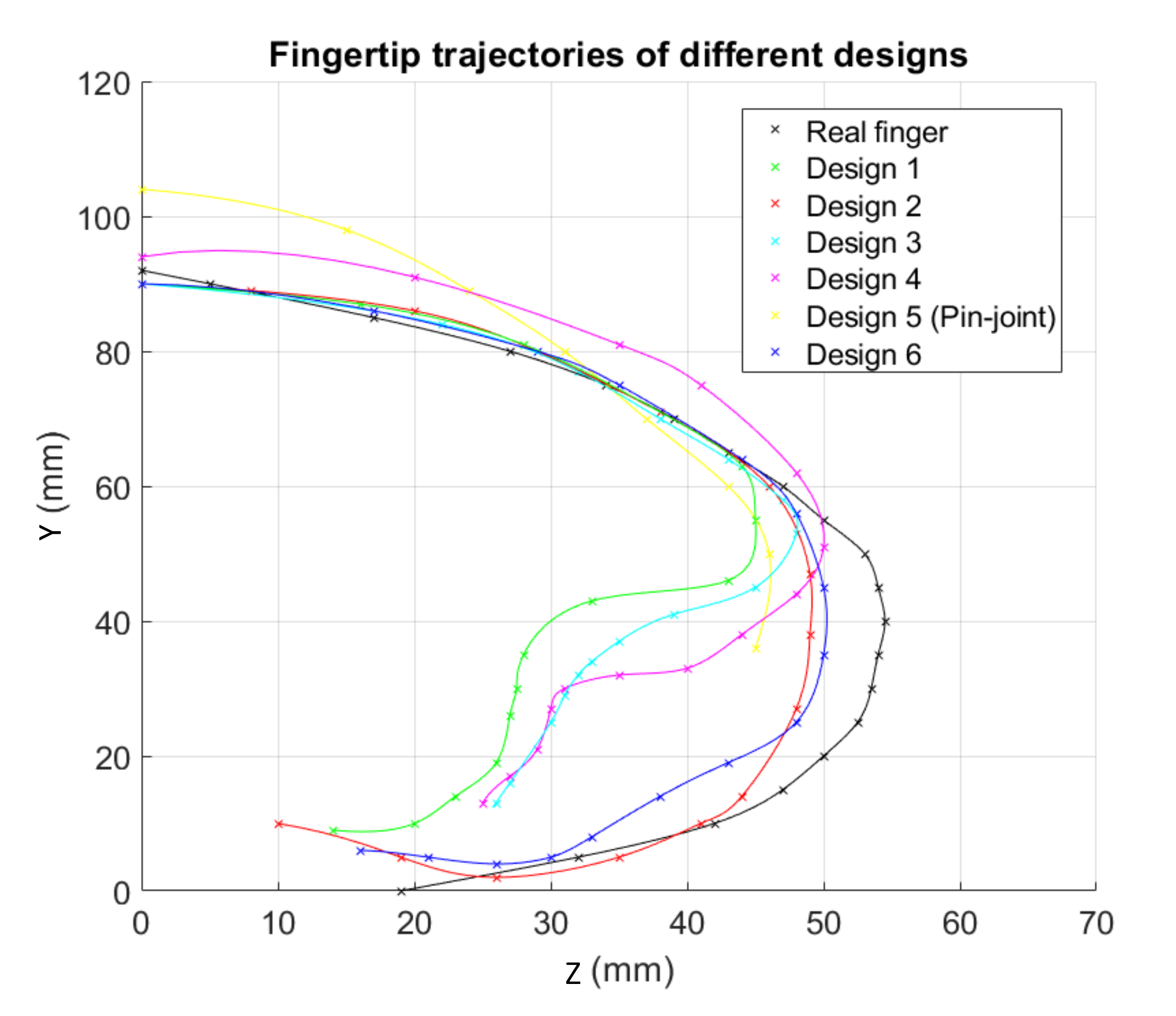}
\caption{Fingertip trajectories during flexion with different designs of the underactuated system. These trajectories are captured from the lateral view. Design 5 is from InMoov hand.}
\label{fig: finger_tip}
\end{figure}

Trajectories from the lateral view (i.e., the $yz$-plane in Fig. \ref{fig: bone}) of the 6 designs are reported in Fig. \ref{fig: finger_tip}. We can observe that the flexion of the baseline (Design 5) is limited, as its minimum $y$ coordinate is larger than that of the human finger by a large margin. On the contrary, all our 5 designs are more flexible and better replicate the human finger. We notice that Design 2 and 6 have the similar performance, therefore we further measure their maximum pull strength to determine the optimal one. With the same servomotor, the maximum pull strength of Design 6 is 2.0 kg, while that of Design 2 is 0.91 kg. Therefore we consider Design 6 as the optimal choice for implementing our underactuated system.

\section{Conclusions}

In this paper, we propose a fast and systematic method to fabricate customizable robotic hands. The core of our method is the skin-tissue-bone, multi-layer design inspired by the anatomy of the human hand. Our fast template matching method can generate the 3D bone models of the target hand, while our concentric tube based structure for the tissue layer ensures that the fabricated hand is of high deformability. With the multi-layer design, we significantly reduce the cost and complexity of the actuation system. Extensive experiments with the standard object grasping types, as well as with special objects, demonstrate that our design is superior to previous robotic hands.

We expect that this study can provide a new baseline for fabricating anthropomorphic robotic hands based on 3D scanning and 3D printing. In the future, we plan to add soft sensors into the hollow chamber of the tissue layer to help to increase the accuracy of object grasping and manipulation. We also plan to devise an actuation system with more degrees of freedom to complete complicated motions besides flexion and extension.

\section{Acknowledgments}
This research is supported by the National Research Foundation, Singapore under its International Research Centres in Singapore Funding Initiative. Any opinions, findings and conclusions or recommendations expressed in this material are those of the author(s) and do not reflect the views of National Research Foundation, Singapore. 

\bibliographystyle{IEEEtran}
\bibliography{references}

\begin{thebibliography}{10}
\providecommand{\url}[1]{#1}
\csname url@samestyle\endcsname
\providecommand{\newblock}{\relax}
\providecommand{\bibinfo}[2]{#2}
\providecommand{\BIBentrySTDinterwordspacing}{\spaceskip=0pt\relax}
\providecommand{\BIBentryALTinterwordstretchfactor}{4}
\providecommand{\BIBentryALTinterwordspacing}{\spaceskip=\fontdimen2\font plus
\BIBentryALTinterwordstretchfactor\fontdimen3\font minus
  \fontdimen4\font\relax}
\providecommand{\BIBforeignlanguage}[2]{{%
\expandafter\ifx\csname l@#1\endcsname\relax
\typeout{** WARNING: IEEEtran.bst: No hyphenation pattern has been}%
\typeout{** loaded for the language `#1'. Using the pattern for}%
\typeout{** the default language instead.}%
\else
\language=\csname l@#1\endcsname
\fi
#2}}
\providecommand{\BIBdecl}{\relax}
\BIBdecl

\bibitem{gama2014anthropomorphic}
E.~N. Gama~Melo, O.~F. Aviles~Sanchez, and D.~Amaya~Hurtado, ``Anthropomorphic
  robotic hands: a review,'' \emph{Ingenier{\'\i}a y Desarrollo}, vol.~32,
  no.~2, pp. 279--313, 2014.

\bibitem{Salisbury1983}
J.~K. Salisbury and B.~Roth, ``Kinematic and force analysis of articulated
  mechanical hands,'' \emph{Journal of Mechanisms, Transmissions, and
  Automation in Design}, vol. 105, no.~1, pp. 35--41, 1983.

\bibitem{rothling2007platform}
F.~Rothling, R.~Haschke, J.~J. Steil \emph{et~al.}, ``Platform portable
  anthropomorphic grasping with the bielefeld 20-dof shadow and 9-dof tum
  hand,'' in \emph{IEEE/RSJ International Conference on Intelligent Robots and
  Systems}.\hskip 1em plus 0.5em minus 0.4em\relax IEEE, 2007, pp. 2951--2956.

\bibitem{schmitz2010design}
A.~Schmitz, U.~Pattacini, F.~Nori \emph{et~al.}, ``Design, realization and
  sensorization of the dexterous icub hand,'' in \emph{IEEE-RAS International
  Conference on Humanoid Robots}.\hskip 1em plus 0.5em minus 0.4em\relax IEEE,
  2010, pp. 186--191.

\bibitem{deimel2016novel}
R.~Deimel and O.~Brock, ``A novel type of compliant and underactuated robotic
  hand for dexterous grasping,'' \emph{The International Journal of Robotics
  Research}, vol.~35, no. 1-3, pp. 161--185, 2016.

\bibitem{controzzi2016sssa}
M.~Controzzi, F.~Clemente, D.~Barone \emph{et~al.}, ``The sssa-myhand: a
  dexterous lightweight myoelectric hand prosthesis,'' \emph{IEEE Transactions
  on Neural Systems and Rehabilitation Engineering}, vol.~25, no.~5, pp.
  459--468, 2016.

\bibitem{piazza2017softhand}
C.~Piazza, M.~G. Catalano, S.~B. Godfrey \emph{et~al.}, ``The softhand pro-h: a
  hybrid body-controlled, electrically powered hand prosthesis for daily living
  and working,'' \emph{IEEE Robotics \& Automation Magazine}, vol.~24, no.~4,
  pp. 87--101, 2017.

\bibitem{liu2020soft}
``Soft humanoid hands with large grasping force enabled by flexible hybrid
  pneumatic actuators.''

\bibitem{piazza2019century}
C.~Piazza, G.~Grioli, M.~Catalano, and A.~Bicchi, ``A century of robotic
  hands,'' \emph{Annual Review of Control, Robotics, and Autonomous Systems},
  vol.~2, pp. 1--32, 2019.

\bibitem{gosselin2008anthropomorphic}
C.~Gosselin, F.~Pelletier, and T.~Laliberte, ``An anthropomorphic underactuated
  robotic hand with 15 dofs and a single actuator,'' in \emph{IEEE
  International Conference on Robotics and Automation}.\hskip 1em plus 0.5em
  minus 0.4em\relax IEEE, 2008, pp. 749--754.

\bibitem{deshpande2011mechanisms}
A.~D. Deshpande, Z.~Xu, M.~J.~V. Weghe \emph{et~al.}, ``Mechanisms of the
  anatomically correct testbed hand,'' \emph{IEEE/ASME Transactions on
  Mechatronics}, vol.~18, no.~1, pp. 238--250, 2011.

\bibitem{xu2016design}
Z.~Xu and E.~Todorov, ``Design of a highly biomimetic anthropomorphic robotic
  hand towards artificial limb regeneration,'' in \emph{IEEE International
  Conference on Robotics and Automation}.\hskip 1em plus 0.5em minus
  0.4em\relax IEEE, 2016, pp. 3485--3492.

\bibitem{faudzi2017index}
A.~A.~M. Faudzi, J.~Ooga, T.~Goto \emph{et~al.}, ``Index finger of a human-like
  robotic hand using thin soft muscles,'' \emph{IEEE Robotics and Automation
  Letters}, vol.~3, no.~1, pp. 92--99, 2017.

\bibitem{tasi2019design}
B.~J. Tasi, M.~Koller, and G.~Cserey, ``Design of the anatomically correct,
  biomechatronic hand,'' \emph{arXiv preprint arXiv:1909.07966}, 2019.

\bibitem{heung2019robotic}
K.~H. Heung, R.~K. Tong, A.~T. Lau \emph{et~al.}, ``Robotic glove with
  soft-elastic composite actuators for assisting activities of daily living,''
  \emph{Soft robotics}, vol.~6, no.~2, pp. 289--304, 2019.

\bibitem{yamaguchi2019human}
T.~Yamaguchi, T.~Kashiwagi, T.~Arie \emph{et~al.}, ``Human-like electronic
  skin-integrated soft robotic hand,'' \emph{Advanced Intelligent Systems},
  vol.~1, no.~2, p. 1900018, 2019.

\bibitem{oh2020untethered}
B.~Oh, Y.-G. Park, H.~Jung \emph{et~al.}, ``Untethered soft robotics with fully
  integrated wireless sensing and actuating systems for somatosensory and
  respiratory functions,'' \emph{Soft Robotics}, 2020.

\bibitem{yang2020hybrid}
Y.~Yang, Y.~Zhang, Z.~Kan, J.~Zeng, and M.~Y. Wang, ``Hybrid jamming for
  bioinspired soft robotic fingers,'' \emph{Soft robotics}, vol.~7, no.~3, pp.
  292--308, 2020.

\bibitem{hashemi2020bone}
S.~Hashemi, D.~Bentivegna, and W.~Durfee, ``Bone-inspired bending soft robot,''
  \emph{Soft Robotics}, 2020.

\bibitem{bainbridge2004berkshire}
W.~S. Bainbridge, \emph{Berkshire encyclopedia of human-computer
  interaction}.\hskip 1em plus 0.5em minus 0.4em\relax Berkshire Publishing
  Group LLC, 2004, vol.~1.

\bibitem{rus2015design}
D.~Rus and M.~T. Tolley, ``Design, fabrication and control of soft robots,''
  \emph{Nature}, vol. 521, no. 7553, pp. 467--475, 2015.

\bibitem{yap2016high}
H.~K. Yap, H.~Y. Ng, and C.-H. Yeow, ``High-force soft printable pneumatics for
  soft robotic applications,'' \emph{Soft Robotics}, vol.~3, no.~3, pp.
  144--158, 2016.

\bibitem{rich2018untethered}
S.~I. Rich, R.~J. Wood, and C.~Majidi, ``Untethered soft robotics,''
  \emph{Nature Electronics}, vol.~1, no.~2, pp. 102--112, 2018.

\bibitem{tian2019design2}
L.~Tian, N.~M. Thalmann, J.~Zheng \emph{et~al.}, ``Design of a highly
  biomimetic and fully-actuated robotic finger,'' in \emph{IEEE Symposium
  Series on Computational Intelligence}.\hskip 1em plus 0.5em minus 0.4em\relax
  IEEE, 2019, pp. 2382--2387.

\bibitem{feix2015grasp}
T.~Feix, J.~Romero, H.-B. Schmiedmayer \emph{et~al.}, ``The grasp taxonomy of
  human grasp types,'' \emph{IEEE Transactions on Human-machine Systems},
  vol.~46, no.~1, pp. 66--77, 2015.

\bibitem{palli2014dexmart}
G.~Palli, C.~Melchiorri, G.~Vassura \emph{et~al.}, ``The dexmart hand:
  Mechatronic design and experimental evaluation of synergy-based control for
  human-like grasping,'' \emph{The International Journal of Robotics Research},
  vol.~33, no.~5, pp. 799--824, 2014.

\bibitem{she2015design}
Y.~She, C.~Li, J.~Cleary \emph{et~al.}, ``Design and fabrication of a soft
  robotic hand with embedded actuators and sensors,'' \emph{Journal of
  Mechanisms and Robotics}, vol.~7, no.~2, 2015.

\bibitem{tian2019design}
L.~Tian, J.~Liu, N.~M. Thalmann \emph{et~al.}, ``Design of a flexible
  articulated robotic hand for a humanoid robot,'' in \emph{IEEE-RAS
  International Conference on Humanoid Robots}.\hskip 1em plus 0.5em minus
  0.4em\relax IEEE, 2019, pp. 572--577.

\bibitem{tian2018methodology}
L.~Tian, N.~Magnenat-Thalmann, D.~Thalmann \emph{et~al.}, ``A methodology to
  model and simulate customized realistic anthropomorphic robotic hands,'' in
  \emph{Proceedings of Computer Graphics International}, 2018, pp. 153--162.

\bibitem{shah2017review}
P.~B. Shah and Y.~Luximon, ``Review on 3d scanners for head and face
  modeling,'' in \emph{International Conference on Digital Human Modeling and
  Applications in Health, Safety, Ergonomics and Risk Management}.\hskip 1em
  plus 0.5em minus 0.4em\relax Springer, 2017, pp. 47--56.

\bibitem{buryanov2010proportions}
A.~Buryanov and V.~Kotiuk, ``Proportions of hand segments,'' \emph{Int. J.
  Morphol}, pp. 755--758, 2010.

\bibitem{li2008validation}
Z.~Li, C.-C. Chang, P.~G. Dempsey \emph{et~al.}, ``Validation of a
  three-dimensional hand scanning and dimension extraction method with
  dimension data,'' \emph{Ergonomics}, vol.~51, no.~11, pp. 1672--1692, 2008.

\bibitem{landsmeer1961studies}
J.~Landsmeer, ``Studies in the anatomy of articulation. i. the equilibrium of
  the" intercalated" bone.'' \emph{Acta Morphologica Neerlando-Scandinavica},
  vol.~3, p. 287, 1961.

\bibitem{buchner1988dynamic}
H.~J. Buchner, M.~J. Hines, and H.~Hemami, ``A dynamic model for finger
  interphalangeal coordination,'' \emph{Journal of biomechanics}, vol.~21,
  no.~6, pp. 459--468, 1988.

\bibitem{brook1995biomechanical}
N.~Brook, J.~Mizrahi, M.~Shoham \emph{et~al.}, ``A biomechanical model of index
  finger dynamics,'' \emph{Medical Nngineering \& Physics}, vol.~17, no.~1, pp.
  54--63, 1995.

\bibitem{langevin2014inmoov}
G.~Langevin, ``Inmoov-open source 3d printed life-size robot,'' \emph{pp. URL:
  http://inmoov. fr, License: http://creativecommons.
  org/licenses/by--nc/3.0/legalcode}, 2014.

\bibitem{shimawaki2007quasi}
S.~Shimawaki and N.~Sakai, ``Quasi-static deformation analysis of a human
  finger using a three-dimensional finite element model constructed from ct
  images,'' \emph{Journal of Environment and Engineering}, vol.~2, no.~1, pp.
  56--63, 2007.

\end{thebibliography}

\end{document}